\documentclass[letterpaper, 10 pt, conference]{ieeeconf}  % Comment this line out if you need a4paper
\IEEEoverridecommandlockouts                              % This command is only needed if 
\usepackage[T1]{fontenc}
\usepackage{graphics} % for pdf, bitmapped graphics files
\usepackage{epsfig} % for postscript graphics files
\usepackage{xcolor,soul}
\usepackage{booktabs}

\title{\LARGE \bf
Grasp Force Assistance via Throttle-based Wrist Angle Control\\ on a Robotic Hand Orthosis for C6-C7 Spinal Cord Injury
}

\author{Joaquin Palacios$^{1\dagger}$, Alexandra Deli-Ivanov$^{1\dagger}$, Ava Chen$^{1\dagger}$,\\ Lauren Winterbottom$^{2}$, Dawn M. Nilsen$^{2\ast}$, Joel Stein$^{2\ast}$, and Matei Ciocarlie$^{1\ast}$% <-this % stops a space
\thanks{This work was supported by the National Institutes of Health: National Institute of Neurological Disorders and Stroke under grant R01NS115652; A.C. is supported by the Eunice Kennedy Shriver National Institute of Child Health and Human Development under award F31HD111301.}
\thanks{\llap{\textsuperscript{1}}{Department of Mechanical Engineering, Columbia University, New York, NY 10027, USA.\texttt{\footnotesize\{jbp2157, ad3805, ava.chen, matei.ciocarlie\}@columbia.edu}}%
}
\thanks{\llap{\textsuperscript{2}}{Department of Rehabilitation and Regenerative Medicine, Columbia University Irving Medical Center, New York, NY 10032, USA.\newline\texttt{\footnotesize\{lbw2136, dmn12, js1165\}@cumc.columbia.edu}}%
}
\thanks{\llap{\textsuperscript{$\dagger$}}{These authors contributed equally to this work.}}% <-this % stops a space
\thanks{\llap{\textsuperscript{$\ast$}}{Co-Principal Investigators}}%
}

\begin{document}

\maketitle
\thispagestyle{empty}
\pagestyle{empty}

%%%%%%%%%%%%%%%%%%%%%%%%%%%%%%%%%%%%%%%%%%%%%%%%%%%%%%%%%%%%%%%%%%%%%%%%%%%%%%%%
\begin{abstract}
Individuals with hand paralysis resulting from C6-C7 spinal cord injuries frequently rely on tenodesis for grasping. However, tenodesis generates limited grasping force and demands constant exertion to maintain a grasp, leading to fatigue and sometimes pain. We introduce the MyHand-SCI, a wearable robot that provides grasping assistance through motorized exotendons. Our user-driven device enables independent, ipsilateral operation via a novel Throttle-based Wrist Angle control method, which allows users to maintain grasps without continued wrist extension. A pilot case study with a person with C6 spinal cord injury shows an improvement in functional grasping and grasping force, as well as a preserved ability to modulate grasping force while using our device, thus improving their ability to manipulate everyday objects. This research is a step towards developing effective and intuitive wearable assistive devices for individuals with spinal cord injury.

%Keywords: Prosthetics~and~Exoskeletons, Physically~Assistive~Devices, Rehabilitation~Robotics\\

\end{abstract}
\vspace{-1mm}

%%%%%%%%%%%%%%%%%%%%%%%%%%%%%%%%%%%%%%%%%%%%%%%%%%%%%%%%%%%%%%%%%%%%%%%%%%%%%%%%
\section{Introduction}
\vspace*{-3pt}

A Spinal Cord Injury (SCI) often results in partial or complete sensorimotor loss in the arms and lower body, leading to reduced functional independence. Restoration of hand function is one of the highest priorities for SCI \mbox{populations \cite{SCI Priorities Review}}. 

Individuals with hand paralysis caused by C6-C7 spinal cord level injuries often have preserved active wrist extension, which allows them to use a compensatory grasping pattern called tenodesis. Tenodesis leverages wrist extension to passively shorten the digit flexor tendons to close the hand, achieving some degree of thumb-finger lateral grip or finger flexion for grasp. However, the grasping forces generated through tenodesis often fall short of what is required for many activities of daily living (ADLs), even after surgical intervention \cite{Tenodesis and WDFHO Forces}. In addition, maintaining a grasp with tenodesis requires constantly exerting effort in wrist extension, which can be uncomfortable or even painful to sustain for prolonged periods.

\begin{figure}[t]
\centering
\includegraphics[trim=0mm 0mm 0.5mm 0mm,clip=true, 
width=0.92\columnwidth]{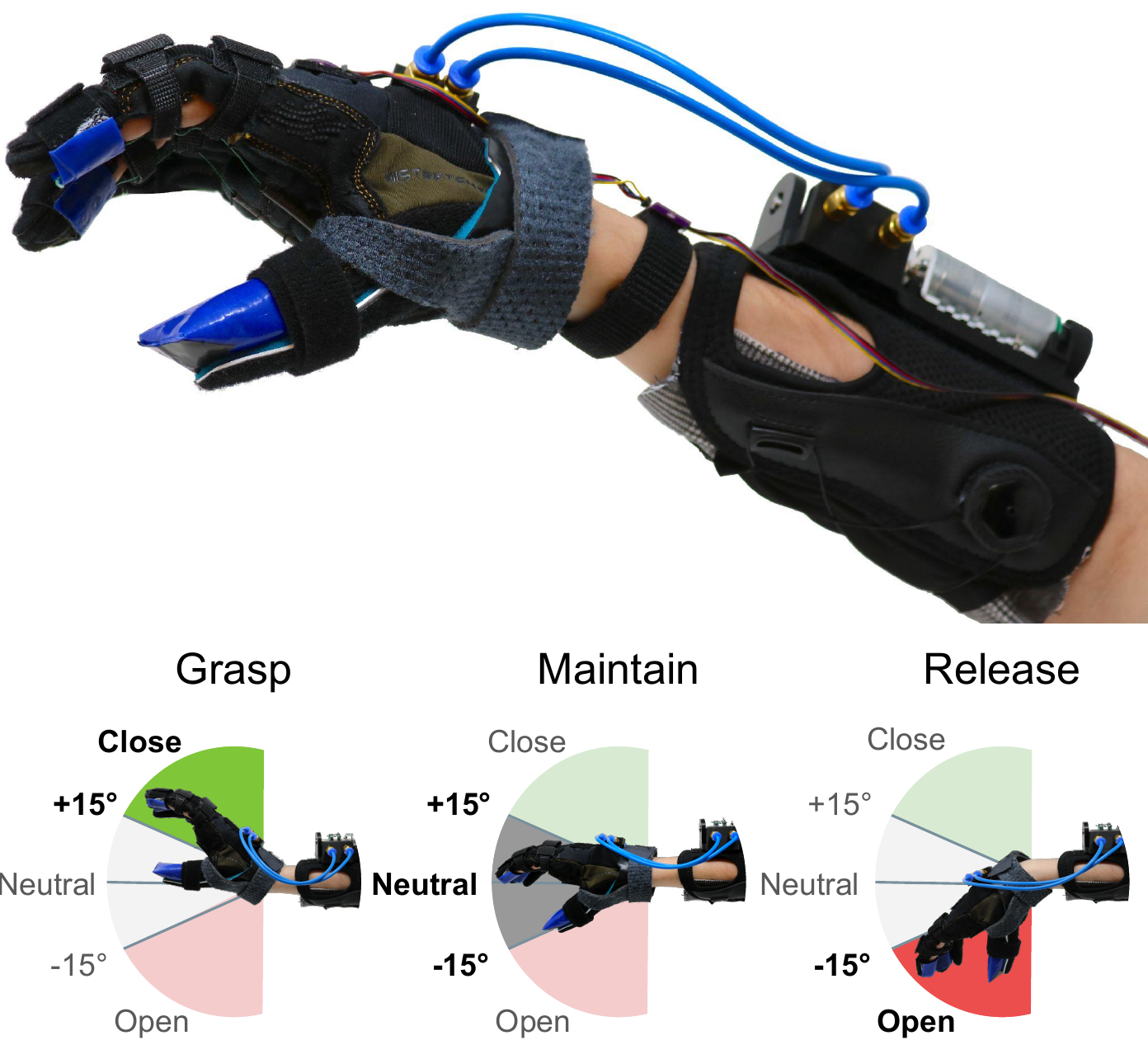}
\vspace{-1mm}
\caption{MyHand-SCI is a robot hand orthosis that is controlled via a Throttle-based Wrist Angle (TWA) control method.}
\vspace{-4mm}
\label{fig: glamour_shot}
\end{figure}

In this paper, we introduce the MyHand-SCI \mbox{(Figure \ref{fig: glamour_shot})}, a robotic orthosis capable of providing active grasping force assistance. Our device is intended for individuals with C6-C7 SCI, and we use wrist angle as a control input, leveraging our target population's familiarity with tenodesis grasping. We explore a \textit{novel control paradigm}, Throttle-based Wrist Angle (TWA) control. In contrast to wrist-based control methods that map wrist angle proportionally to grasp aperture, TWA uses wrist angle thresholds to distinguish between “active” regions, which continuously increase or decrease grasping aperture and thereby force, and a “passive” region, which maintains the current grasp at a neutral wrist angle. 

Through a pilot case study, we show this assistive device improves the functional grasping performance and grasping strength of a research participant with C6 SCI, and demonstrate that TWA control enables the user to intuitively and independently operate the device without requiring continual wrist exertion, reducing fatigue and enabling prolonged use. Moreover, we demonstrate that TWA enables fine motor control in the form of \textit{grasping force modulation}, which is critical to being able to safely operate assistive devices, manipulate delicate objects, and adapt to objects of different shapes, sizes, and deformability. Overall, the main contributions of this work are:
\begin{itemize}
    \item We introduce Throttle-based Wrist Angle (TWA) user control for a grasping assistive device, which allows the user to modulate or hold grasping force using only moderate wrist extension, and without the need to maintain wrist extension for the duration of the grasp.
    \item We use TWA on a novel wearable assistive orthosis and study its performance in both functional tasks and grasp force assistance and modulation. To the best of our knowledge, this is the first example of ipsilateral user-controlled grip force modulation with a wearable robotic orthosis for SCI. 
\end{itemize}

\section{Related Work}

A variety of wearable assistive devices have been developed to assist in grasping for individuals with SCI \cite{Review Paper}, primarily by amplifying grasping strength, but independent user operation of these devices remains a challenge. Many are successful in aiding grasping but use button control, which poses difficulty for a bilaterally-impaired user. Passive wearable devices like Wrist-Driven Flexor Hinge Orthotics (WDFHOs) present a purely mechanical alternative, which takes advantage of tenodesis as a user control and amplifies grasping force using linkages. WDFHOs leverage the intuitive connection between wrist flexion and grasping this population is accustomed to, making them easy to operate independently and allowing for ipsilateral control. However, WDFHOs often still fall short of the forces required for many ADLs \cite{Tenodesis and WDFHO Forces} since they are restricted in the mechanical advantage they can produce from the wrist's limited strength and range of motion.  Additionally, these devices require the user to continually exert their wrist to maintain a desired grasp, which can lead to fatigue and prohibit prolonged use. 

Some active assistive devices have taken a similar approach to user control as WDFHOs by using the wrist as a control input \cite{Tenodesis-Grasp-Emulator, Hybrid Orthosis, Exo-Glove, chang2023}. Tenodesis is leveraged as an intuitive and familiar motion, using the wrist as an ipsilateral control input, but actuators are added to decouple the wrist from force generation, generating more suitable grasping forces. Wrist-controlled assistive devices fall into two categories we call Binary Wrist Angle (BWA) control and Proportional Wrist Angle (PWA) control. Devices that use BWA control \cite{Hybrid Orthosis, Exo-Glove, yoo2019} treat grasping as a binary where the hand can be either relaxed or grasping and use a wrist angle threshold to activate each mode. This simplified grasping binary limits how much the user can adapt to objects of different sizes or modulate grasp force for fragile objects. Other devices \cite{chang2023} employ PWA control, in which the user's wrist angle is mapped proportionally to the degree of grasping, similar to WDFHOs. This offers more control over how the user grasps objects, but it is still constrained by the wrist's limited range of motion. Moreover, both BWA and PWA control, like WDFHOs, require the user to maintain wrist extension to hold a desired grasp, making them physically tiring. 

\section{MyHand-SCI Assistive Device Design}

\begin{figure}[t]
    \centering
    \includegraphics[width=\columnwidth]{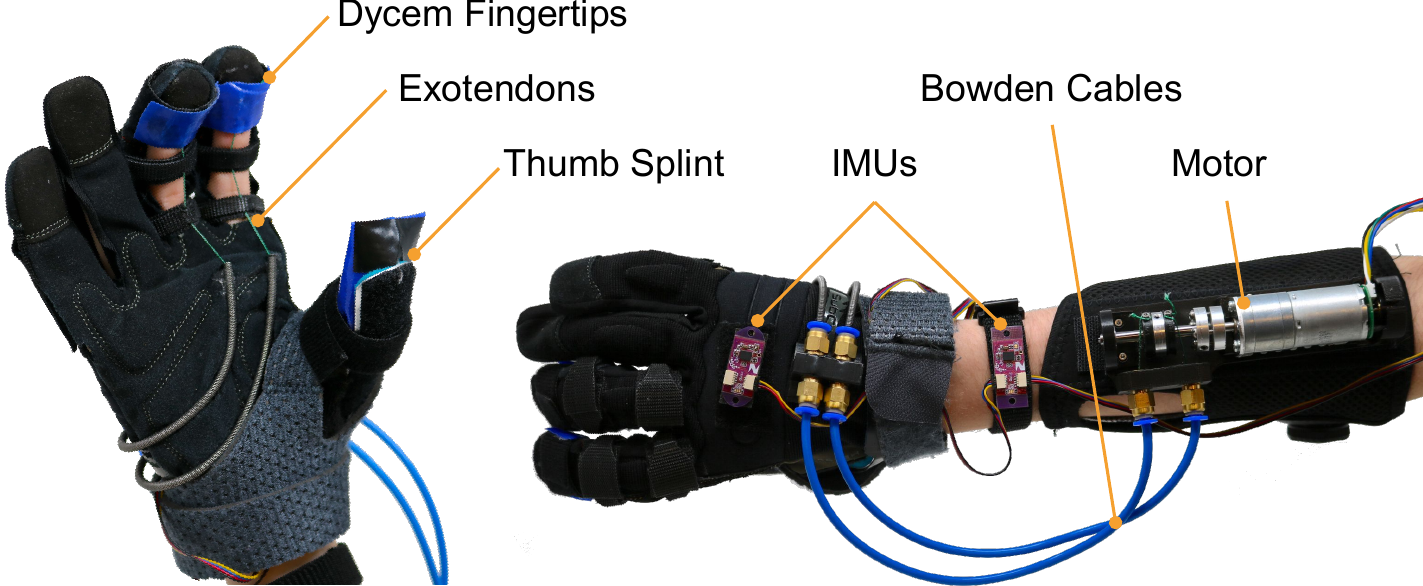}
    \caption{Diagram of MyHand-SCI assistive device with labeled components.}
    \label{fig: device_diagram}
\end{figure}

The MyHand-SCI is a wrist-controlled, exotendon-driven wearable assistive orthosis (Figure \ref{fig: device_diagram}) intended to assist pinch and power grasps, which are versatile grasps in terms of the sizes and types of objects a user can grab and utilize for ADLs. To allow for tenodesis-inspired wrist user control, leaving the wrist's motion unencumbered was an important requirement. We use a tendon-driven design to transfer force from a motor mounted on a forearm brace to the fingers without affecting the wrist. The forearm brace terminates before the wrist joint to enable free mobility and Bowden cables are used to bypass the wrist and transmit force from the motor to the fingers with minimal friction. Notably, unlike linkage-based orthoses, our exotendon and Bowden cable design does not constrain the wrist's range of motion or kinematics in any axis.  

\begin{figure}[]
    \centering
    \includegraphics[width=\columnwidth]{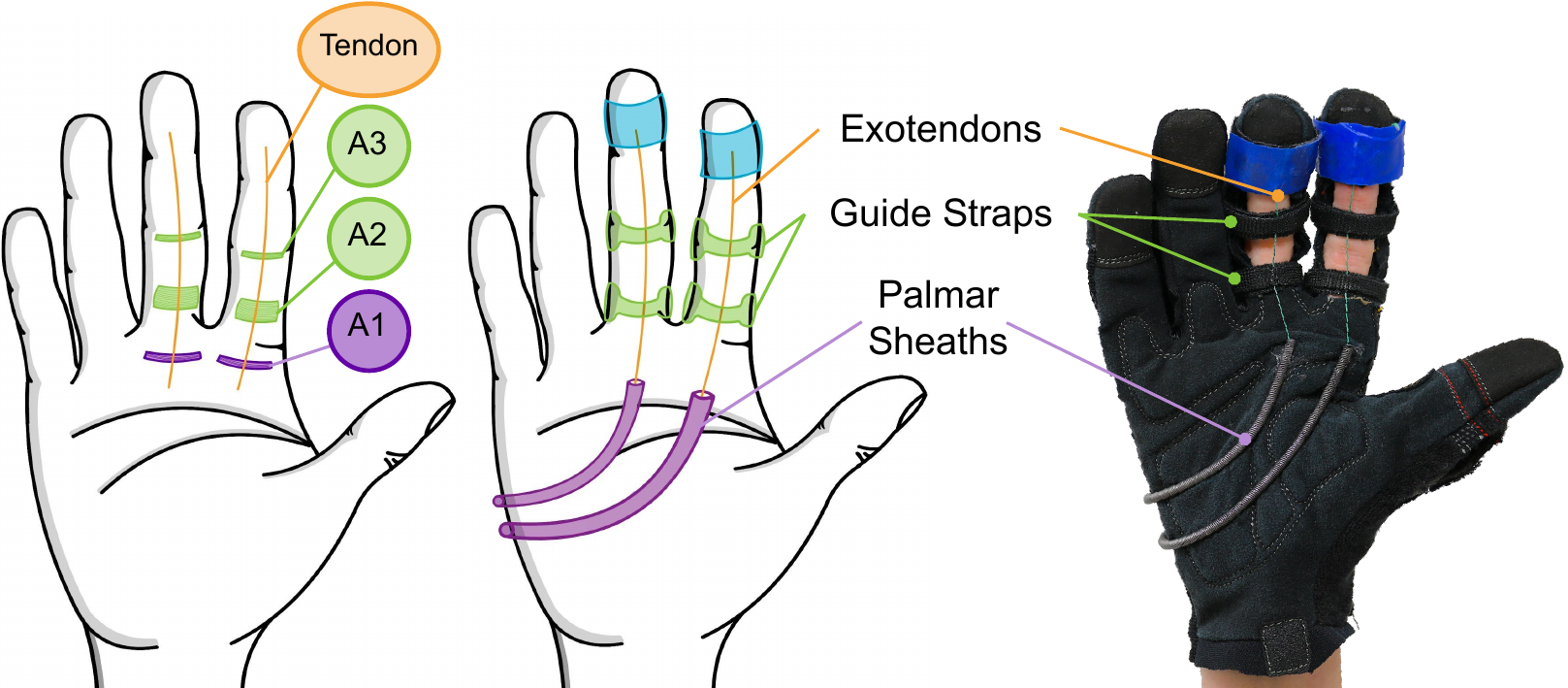}
    \caption{Bioinspired tendon routing. Left: Simplified diagram with anatomical annular ligaments (tendon pulleys). Middle: Device diagram with anatomical analogues. Right: Device palmar side.}
    \label{fig: routing_diagram}
    \vspace{-3mm}
\end{figure}

This prototype actuates digits 2 and 3 via exotendons routed from each finger to a single motor mounted on a forearm brace (Figure \ref{fig: device_diagram}); digits 4 and 5 are unassisted. From the palm onwards, the exotendons follow a bioinspired routing that mimics the anatomy of the finger by configuring guide straps to roughly coincide with the location of the anatomical A1, A2, and A3 annular ligaments (Figure \ref{fig: routing_diagram}). These straps act as tendon pulleys that encourage natural finger flexion. Dycem padding at the fingertips improves friction for a more secure grip. The thumb is stabilized in opposition with a static fabric and aluminum splint to support pinch and power grasps. The overall device weighs 387 grams (excluding the tabletop controller and power supply).

\section{Throttle-based Wrist Angle Control} 

\begin{figure*}[t!]
    \centering
    \includegraphics[trim=0mm 0mm 0mm 0mm,clip=true, width=0.85\textwidth] {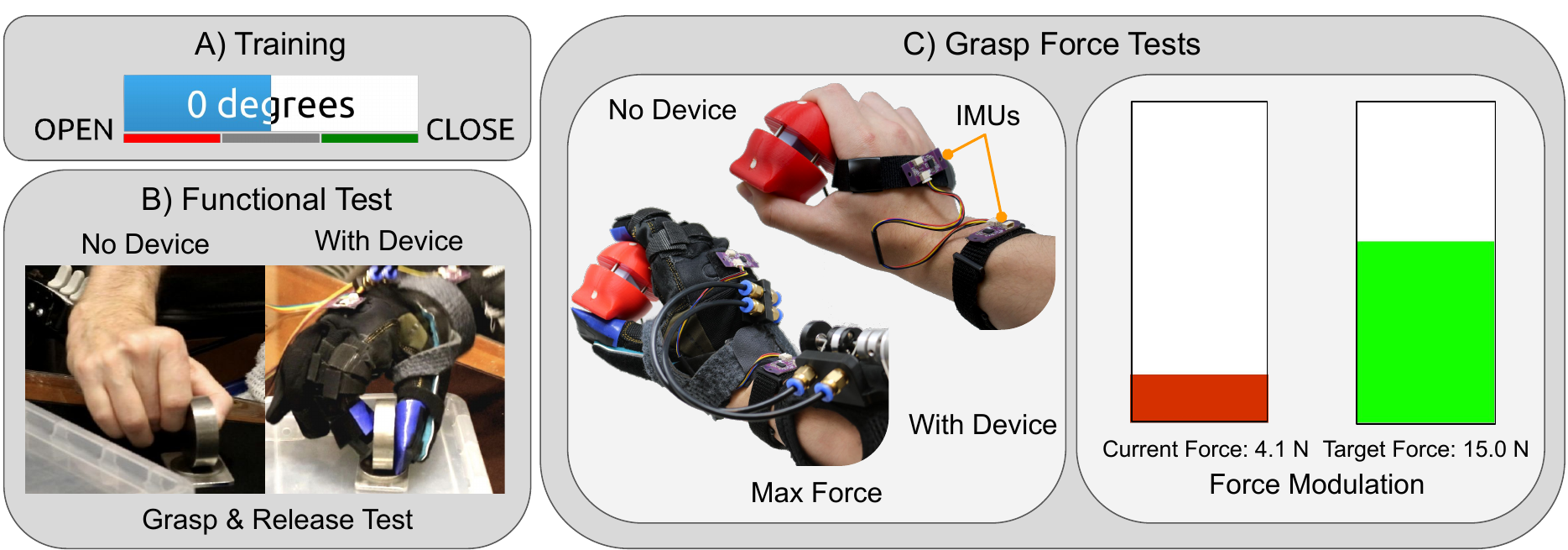}
    \caption{Experimental Protocol.}
    \vspace{-6mm}
    \label{fig: experiment_diagram}
\end{figure*}

The MyHand-SCI aims to leverage our target population's residual motor skills by using the familiar motion of tenodesis as a user control, but also allow the user to maintain a grasp without the need for continual wrist extension. To that end, we introduce a novel wrist-driven user control method, Throttle-based Wrist Angle (TWA) control, which uses wrist angle as input and threshold values to differentiate between three regions (Figure \ref{fig: glamour_shot}) as follows:
\begin{itemize}
    \item An input angle in the “Close” region drives the motor to pull on the exotendons, continuously increasing flexion of the user's fingers or increasing grasp force.
    \item An angle in the “Neutral” region maintains the current motor position.
    \item An angle in the “Open” region drives the motor in the opposite direction, releasing tension from the exotendons and allowing the user to extend their fingers or reduce grasp force.
\end{itemize}

When in an active region, like “Close” and “Open”, the device will continue to actuate the motor in the desired direction until either the user moves the wrist into the “Neutral” region, or a safety limit is hit. For example, in order to achieve a desired level of grasp force, the user simply needs to keep their wrist in the “Close” region, thereby allowing the motor to pull until they are satisfied with the force level, and then move to the “Neutral” region to preserve the current level of force. This also means that the user is not restricted by wrist range of motion but instead can access any degree of finger flexion with a moderate angle of wrist extension. 

Unlike PWA control, TWA control allows the user to achieve a desired grasp and then return to the “Neutral” region and maintain the grasp, without the need for continual exertion. Additionally, in contrast to BWA this control method is not restricted to a single grasping force or configuration. This simple control method takes advantage of our target population's proficiency using tenodesis grasping, allowing them to self-modulate the degree of finger flexion and therefore the grasping force applied.

The orthosis is equipped with two inertial measurement units (IMUs), one on a wrist bracelet and another on the back of the hand, as seen in Figure \ref{fig: device_diagram}, which are used to determine the user's angle of wrist extension/flexion. We empirically found ±15 degrees to be comfortable thresholds for the “Neutral” region to be large enough to not accidentally trigger the “Open”/“Close”, but small enough so that subtle wrist motions could operate the device.

\section{Experiments}

For initial validation of our device and control method, we conducted a series of experiments with a participant with C6 tetraplegia who provided informed consent to participate in the study. Testing took place at Columbia University Irving Medical Center in accordance with the protocol (IRB-AAAU2339) approved by the Columbia University Institutional Review Board. This participant had prior experience with earlier versions of the robotic device, TWA control method, and experimental protocol. All experiments were performed under the supervision of an occupational therapist.

To quantify the success of our design objectives described in Sections III and IV, we devised an experimental protocol consisting of a Functional Test and two Grasping Force Tests. All tests were performed first without wearing the device to obtain a baseline and then again with the assistive device operated independently by the research participant through TWA user control. The breakdown of these experiments can be seen in Figure \ref{fig: experiment_diagram}, and each experiment phase will be explained in the following sections. 

Before tests involving using the assistive device, the participant was given a training period to get accustomed to TWA control and grasping assistance from the MyHand-SCI. The first training was performed with the motor disconnected and involved showing the participant a graphical user interface (Figure \ref{fig: experiment_diagram}.A) displaying their current wrist angle measurement and the thresholds that would activate the device. This served two purposes, getting the user trained for TWA, and calibrating for appropriate wrist angle thresholds, for which they found $\pm$ 15 degrees to be comfortable. A second training was performed with the motor connected, where the participant was allowed to operate the device, first with a cautious motor position safety limit and velocity. These settings were progressively increased as the participant grew more comfortable with the device and control method.  

\subsection{Functional Performance Test}

The primary objective of this assistive device is functional performance, that is, to improve the user's ability to independently perform ADLs. To evaluate this we selected the standardized Grasp and Release Test (GRT), which assesses the ability of an individual with C5-C7 tetraplegia to manipulate a set of objects varying in size and weight (Figure \ref{fig: experiment_diagram}.B). The GRT is scored per object by the number of successes achieved within 30 seconds. The timing component of this test also allows us to assess the user's comfort operating the device independently and the responsiveness of TWA control. As per GRT instructions, the participant was allowed to practice grasping the objects both with and without the device before their respective tests.
\vspace{-1mm}

\subsection{Maximum Grasping Force Test}

\begin{figure}[t]
    \centering
    \includegraphics[trim=0mm 0mm 0mm 5mm,clip=true,width=1\columnwidth]{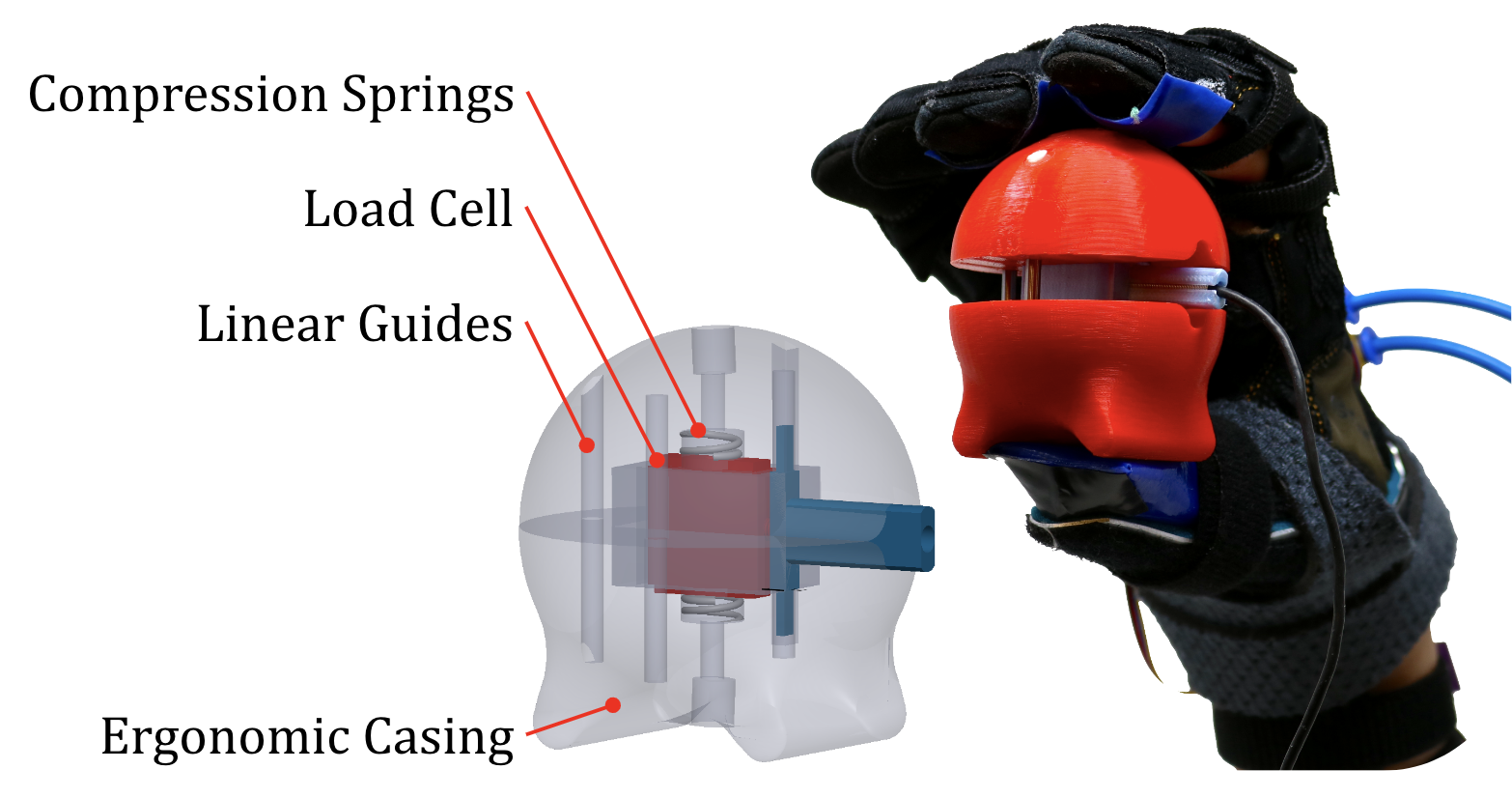}
    \vspace{-7mm}
    \caption{Instrumented object for measuring grasping force.}
    \label{fig: force_apparatus}
    \vspace{-6mm}
\end{figure}

In order to measure grasp forces during our experiments, we developed an instrumented object housing a Futek FSH00097 uniaxial load cell (sensor resolution 0.28N) encased in an ergonomic 3D printed shell, as seen in Figure \ref{fig: force_apparatus}. The instrument contains compression springs in series with the load cell to provide tactile force feedback to the participant, and linear guides ensure only uniaxial forces are applied to the sensor. 

To assess grasping force performance, the first test conducted was a maximum force test (Figure \ref{fig: experiment_diagram}.C) to determine how much the device augments the user's grasping force. To obtain a baseline maximum force, the instrumented object was placed in the participant's hand and they were asked to squeeze as hard as possible without over straining their hand. The same procedure was repeated with the device, this time with TWA-controlled active assistance. For both the baseline and assisted max force tests the participant performed three trials, and the highest force value among them was recorded. 

\subsection{Grasping Force Modulation Test}

The second grasping force test aimed to assess grasping force modulation, i.e. the user's ability to accurately control the magnitude of their grasping force. Like the maximum force test, this test was performed both with and without device assistance, to determine if the participant could modulate force and if so compare device performance to this baseline.

To evaluate grasping force modulation we chose different grasping force targets that the participant would attempt to match using the instrumented object. We calculated the target forces as percentages of the recorded max force both with and without device assistance, and the targets consisted of a low force target (20\% of max force), a medium force target (50\% of max force), and a high force target (80\% of max force), for a total of 6 targets. For each target, the participant was shown a graphical display with the target force and a live force reading from the instrumented object, each depicted as a vertical bar as shown in  Figure \ref{fig: experiment_diagram}.C. 

The participant was directed to grip the instrumented object and modulate their grasping force until it matched the target or they became tired. A target band of ±1N was used as an acceptable range for the grasp force to be considered sufficiently close to the target, and the graphical display indicated when the force was within this bound by turning a bright green color. If the participant was able to reach this, they were instructed to maintain their grasp force within the target band for 3 seconds if possible.

Three trials were conducted per target force, and each trial concluded when either the participant maintained the force within the target band for 3 seconds, which would be considered a “success”, or if 30 seconds elapsed, in which case the trial would be considered “unsuccessful”. A “modulation time”, defined as the time until the target was reached and then maintained for 3 seconds, was recorded for each trial. For all grasping force performance tests without the device, the participant was outfitted with IMUs in the same positions as in the MyHand-SCI, as seen in \mbox{Figure \ref{fig: experiment_diagram}.C}, to obtain wrist angle measurements during unassisted tenodesis grasping.

\vspace{-1mm}

\section{Results}

\subsection{Functional Performance}

\begin{figure}[b]
    \centering
    \includegraphics[trim=2mm 1mm 1mm 1mm,clip=true,width=0.99\columnwidth]{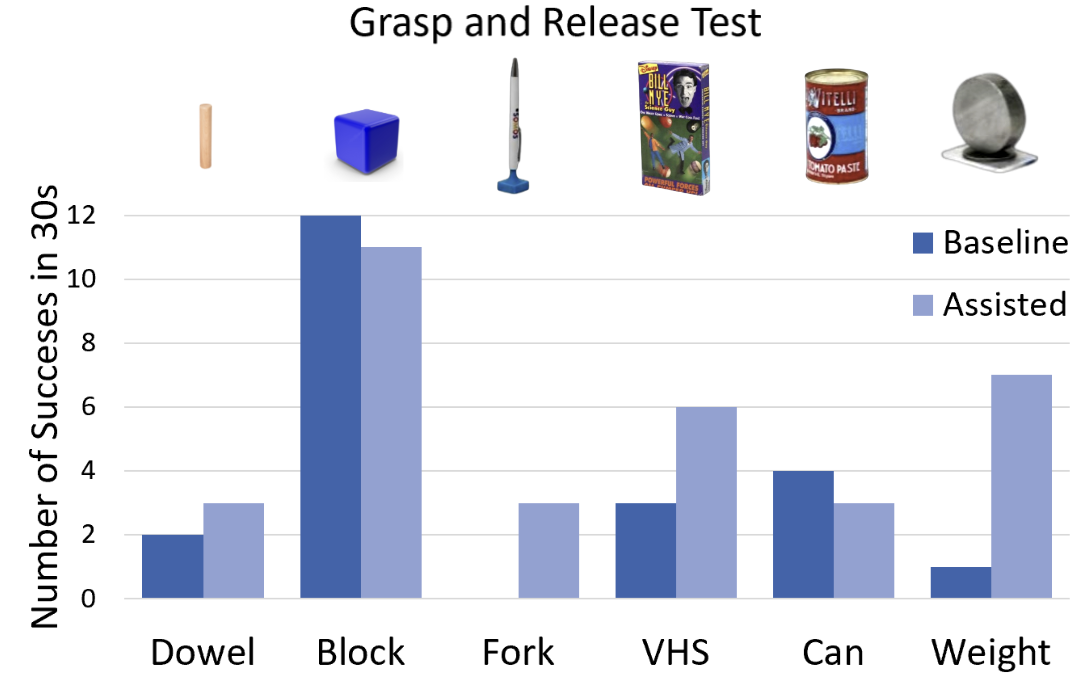}
    \caption{Grasp and Release Test scores for research participant with C6 tetraplegia, with and without device assistance. Objects are ordered in increasing weight from left to right.}
    \label{fig: grt_results}
\end{figure}

Within one session, the participant obtained a GRT score of 33 with the assistive device compared to 22 without it. The scores reflect the cumulative amount of times the participant successfully grasped and released the items, given a 30-second interval for each object. Figure \ref{fig: grt_results} shows specific performance per item. The participant responded well to the device, showing no signs of discomfort or adverse effects, and was able to determine and modulate (through tenodesis control) how much force he needed from the device. The participant's wrist mobility and range of motion were undisturbed by the device. Moreover, the participant described his experience using the device and TWA control as “intuitive” and commented, “It's the way we usually grasp using tenodesis”. 
\vspace{-4mm}

\vspace{3mm}
\subsection{Maximum Grasping Force}

\begin{table*}[t!]
    \renewcommand{\arraystretch}{1}
    \caption{Max force and force modulation results.}
    \label{tab: force_results}
    \centering
    \begin{tabular}{c|cc|cccc}
    \midrule
    & \multicolumn{2}{c}{\textbf{Max Force Test}} & \multicolumn{4}{c}{\textbf{Force Modulation Test}} \\ 
    \midrule
    \textbf{} & 
    \begin{tabular}[c]{@{}c@{}}\textbf{Average}\\\textbf{Max Force (N)} \end{tabular} & 
    \begin{tabular}[c]{@{}c@{}}\textbf{Highest}\\\textbf{Max Force (N)} \end{tabular} & 
    \begin{tabular}[c]{@{}c@{}}\textbf{Target}\\\textbf{(\% of Highest Force)} \end{tabular} &
    \textbf{Target Force (N)} &
    \begin{tabular}[c]{@{}c@{}}\textbf{Average}\\\textbf{Modulation Time (s)} \end{tabular} &
    \textbf{\# of Successes} \\ 
    \midrule
    \textbf{No Device} & $8.4 $ & $10.5 $& 20\% & 2.1 & $3.00$&  3 \\ 
    & & & 50\% & 5.3 & $16.23$& 3 \\ 
    & & & 80\% & 8.4 & $5.78$&  3 \\ 
    \midrule
    \textbf{With Device} & $13.7 $ & $15.3 $& 20\% & 3.1 & $17.11$& 2 \\ 
    & & & 50\% & 7.7 & $13.43$&  3 \\ 
    & & & 80\% & 12.2 & $10.53$&  3 \\ 
    \bottomrule
    \end{tabular}
    \vspace{2mm}
\end{table*}

\begin{figure}[t]
    \centering
    \includegraphics[width=1.0\columnwidth]{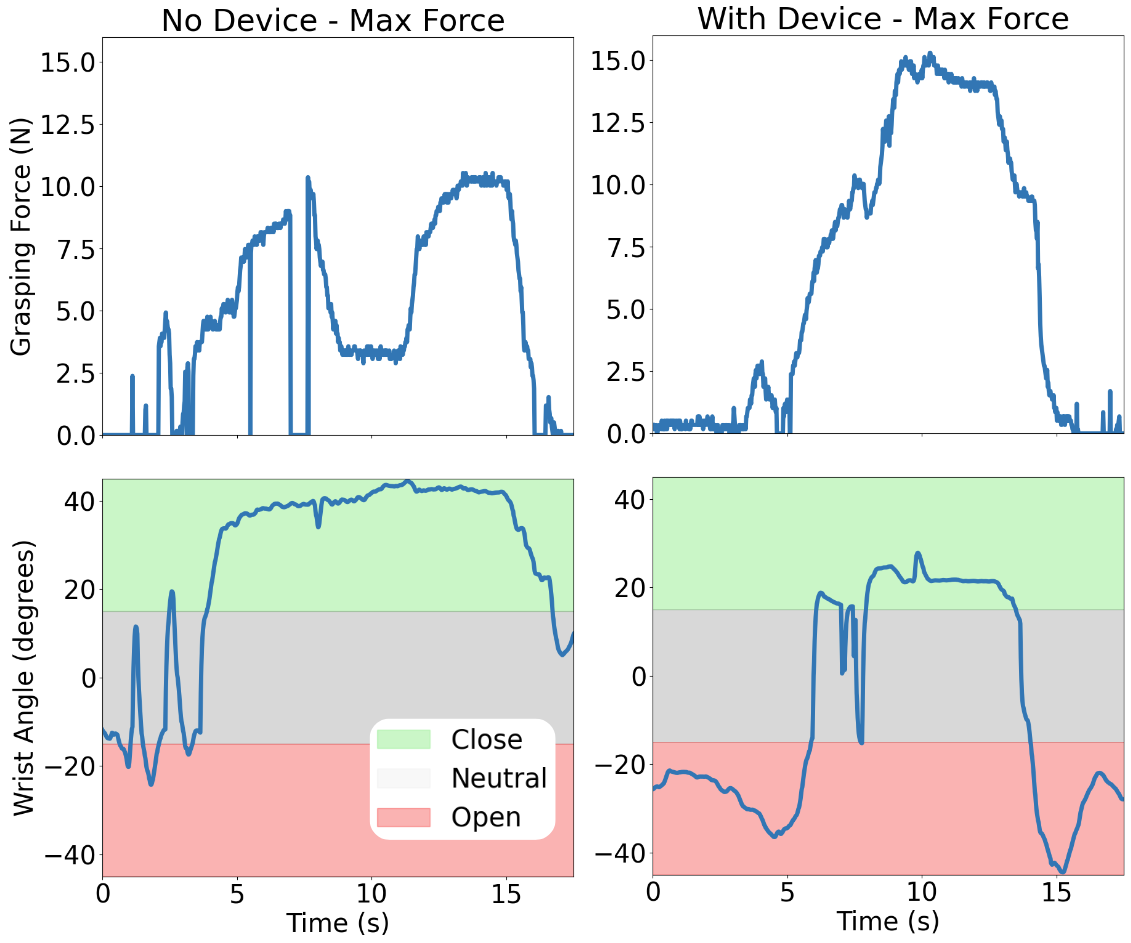}
    \caption{Comparison between Highest Max Force attempts with and without the device. The participant has to extend their wrist considerably less using the device ($\sim$25° compared to $\sim$40°) and achieves a higher force.}
    \label{fig: amax_force}
    \vspace{4mm}
\end{figure}

Within one session, the participant was able to generate a 63.1\% increase in average maximum grasping force with device assistance, reaching an average of 13.7N throughout the three max force trials with active assistance compared to an average of 8.4N without it, as noted in Table \ref{tab: force_results}. Figure \ref{fig: amax_force} shows the maximum force trials in which the participant achieved the highest grasping forces, which we used to calculate the target forces. Notably, during this trial the device allowed the participant to extend their wrist less (by about 15 degrees) while achieving a higher max force (Figure \ref{fig: amax_force}, Right), compared to the highest max force attempt without the device (Figure \ref{fig: amax_force}, Left). 

\subsection{Grasp Force Modulation}

\begin{figure}[t]
    \centering
    \includegraphics[width=\columnwidth]{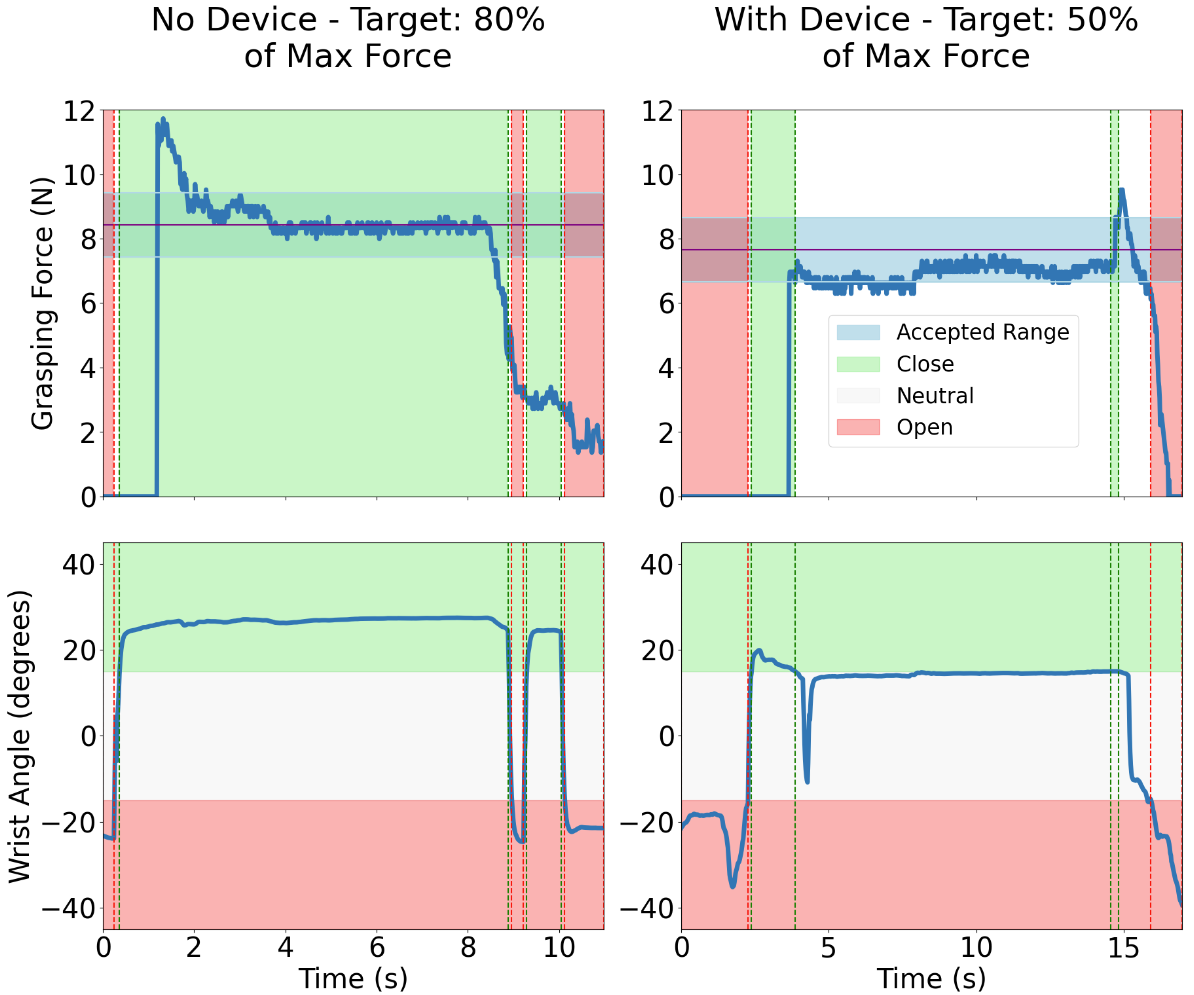}
    \caption{Force modulation results visualized for similar magnitude targets with and without device assistance. With the device, grasping force can be maintained without continual wrist exertion, and to achieve similar force without assistance the patient has to extend their wrist more (28+ degrees compared to 20 degrees).}
    \label{fig: angle_force}
\end{figure}

Table \ref{tab: force_results} also shows the participant was able to modulate force both with and without the device, with only one unsuccessful device-assisted attempt. This table also includes the average modulation time (defined in Section IV) for each target. Lastly, Figure \ref{fig: angle_force} displays a representative pair of successful trials with and without the device. As seen in this time-series, the user can reach a desired force and then return to a neutral wrist angle, without requiring constant wrist extension to maintain a desired grasping force.

\section{Discussion}

Our GRT assessment (Figure \ref{fig: grt_results}) shows that the MyHand-SCI provided functional assistance and did not hinder the participant’s baseline hand function as the results were mostly on par with or greater than the baseline. What we noticed is that for objects that require minimal forces to lift, the participant did not need to use the device to complete the GRT assessment, instead relying on their current abilities. This reveals how the device does not get in the way of the user, and instead only provides as much assistance as is needed, either actuating minimally or fully depending on the inputs from the user. Device assistance was particularly helpful for heavier objects that required a higher grasping force to lift successfully. The device made it possible for the participant to manipulate GRT objects that the user was unable to or struggled with previously, like the fork, while also relieving strain from the hand when maintaining grasps or trying to lift heavier objects, like the VHS (Figure \ref{fig: grt_results}). This provides initial validation for our mechanical design, control method, and generally the orthosis' effectiveness as an assistive device. The only exception was the can, with which the participant had some difficulty due to its larger size which requires greater thumb opposition and abduction. Based on this result, we determined thumb placement is a limitation in our design that merits further prototyping. This was exacerbated by the fact our device does not assist finger extension, which limited grip aperture and posed a difficulty for the participant when grasping large objects. A future prototype of the device should implement assisted extension. Nonetheless, overall the device showed promising results in improving functional grasping performance. 

Our study's research participant had remarkably good baseline performance, both in terms of maximum grasping force and force modulation. This was supported by the fact the participant reported regular use of the right hand in daily activities such as self-feeding using a universal cuff and having undergone tendon transfer surgery to improve hand function. Even compared to this strong baseline, our device still enhanced grasping assistance to a degree that improved functional performance, as evidenced by the GRT results. Future work will expand the participant pool and include individuals with different degrees of impairment.

Table \ref{tab: force_results} shows the MyHand-SCI allows for successful force modulation, especially at higher target forces. However, we also note that our device increased Average Modulation Time, which suggests the device control could be further tuned to the individual, particularly in terms of motor velocity and angle thresholds. This effect was stronger at low forces, where we qualitatively observed significant overshoot when performing assisted modulation. We believe that future work can address this issue via better tuning of the motor speed to the user's abilities and needs. 

We also note that the instrumented object used in this study to measure grasp force uses a single axis sensor and thus only records grasping forces along the main direction of the grasp, normal to the object surface. It is possible that in the course of the experiment, SCI users will be able to generate tangential, frictional forces as well. Future studies should thus consider a version of the measurement device instrumented to record off-axis forces.

\section{Conclusion}

We have introduced the MyHand-SCI, a wearable assistive robot that provides functional grasping assistance and force modulation. Our device puts forward a novel user control paradigm dubbed TWA control, which is based on the natural tenodesis movements that individuals with C6-C7 SCI are accustomed to. In contrast to binary control methods, TWA control offers the user finer control of the device, allowing them to modulate grasping force and adapt to objects of varying sizes. Additionally, unlike proportional control methods, TWA allows the user to operate the device with moderate degrees of wrist extension or flexion respectively and maintain grasps even at a neutral, comfortable wrist angle. Based on our pilot study with a participant with C6 tetraplegia, TWA control shows promising potential as a form of ipsilateral user control, enabling the user to perform motor tasks requiring force modulation. Future work can delve into 1) quantitative comparisons between TWA and alternative user control methods such as BWA or PWA, 2) the performance of our approach for various levels of hand impairment, aiming to provide assistance to a wider range of users, and 3) the impact of fine-tuning control settings for personalized assistance as well as a longer training period on the MyHand-SCI's usability and performance.

\section*{ACKNOWLEDGMENTS}

The authors would like to thank Grace Munger, Katelyn Lee, Katherine O'Reilly, and Carolyn David for assistance with experiment setup.


%%%%%%%%%%%%%%%%%%%%%%%%%%%%%%%%%%%%%%%%%%%%%%%%%%%%%%%%%%%%%%%%%%%%%%%%%%%%%%%%

\end{document}